\documentclass{llncs}

\usepackage{times}

\usepackage{latexsym}
 
\usepackage{graphicx}
\usepackage{wrapfig}

\usepackage{xspace}

\begin{document}

\title{CSPLib: Twenty Years On\thanks{An extended version of the original CP-1999 paper about CSPLib is available without pay-wall at https://tinyurl.com/CSPLibAt20}
}

\author{Ian P. Gent\inst{1} \and Toby Walsh\inst{2}}
\institute{University of St Andrews\\
Scotland\\
Ian.Gent@st-andrews.ac.uk
\and
TU Berlin and UNSW Sydney\\
Germany and Australia\\
tw@cse.unsw.edu.au}

\maketitle

\begin{abstract}
In 1999, we introduced CSPLib, a benchmark library for the constraints
community. 
Our CP-1999 poster paper about CSPLib \cite{DBLP:conf/cp/GentW99} discussed the advantages
and disadvantages of building such a library. Unlike some other
domains such as theorem proving, or machine learning,
representation was then and remains today a major issue in the success or failure to
solve problems. Benchmarks in
CSPLib are therefore specified in natural language as this allows
users to find good representations for themselves. 
The community responded positively and CSPLib has become
a valuable resource but, as we discuss here, 
we cannot rest.
\end{abstract}

\newcommand{\DLex}{\mbox{\sc DoubleLex}}
\newcommand{\snakelex}{\mbox{\sc SnakeLex}}

\newcommand{\set}{\mathcal}
\newcommand{\myset}[1]{\ensuremath{\mathcal #1}}

\renewcommand{\theenumii}{\alph{enumii}}
\renewcommand{\theenumiii}{\roman{enumiii}}
\newcommand{\figref}[1]{Figure \ref{#1}}
\newcommand{\tref}[1]{Table \ref{#1}}
\newcommand{\myldots}{\ldots}

\newtheorem{myproblem}{Problem}
\newtheorem{mydefinition}{Definition}
\newtheorem{mytheorem}{Proposition}
\newtheorem{mylemma}{Lemma}
\newtheorem{myexample}{Running Example}{\bf}{\it}
\newtheorem{mytheorem1}{Theorem}
\newcommand{\myproof}{\noindent {\bf Proof:\ \ }}
\newcommand{\myqed}{\mbox{$\Box$}}
\newcommand{\myend}{\mbox{$\clubsuit$}}

\newcommand{\mymod}{\mbox{\rm mod}}
\newcommand{\mymin}{\mbox{\rm min}}
\newcommand{\mymax}{\mbox{\rm max}}
\newcommand{\range}{\mbox{\sc Range}}
\newcommand{\roots}{\mbox{\sc Roots}}
\newcommand{\myiff}{\mbox{\rm iff}}
\newcommand{\alldifferent}{\mbox{\sc AllDifferent}}
\newcommand{\permutation}{\mbox{\sc Permutation}}
\newcommand{\disjoint}{\mbox{\sc Disjoint}}
\newcommand{\cardpath}{\mbox{\sc CardPath}}
\newcommand{\CARDPATH}{\mbox{\sc CardPath}}
\newcommand{\common}{\mbox{\sc Common}}
\newcommand{\uses}{\mbox{\sc Uses}}
\newcommand{\lex}{\mbox{\sc Lex}}
\newcommand{\usedby}{\mbox{\sc UsedBy}}
\newcommand{\nvalue}{\mbox{\sc NValue}}
\newcommand{\slide}{\mbox{\sc CardPath}}
\newcommand{\sliden}{\mbox{\sc AllPath}}
\newcommand{\SLIDE}{\mbox{\sc CardPath}}
\newcommand{\circularslide}{\mbox{\sc CardPath}_{\rm O}}
\newcommand{\among}{\mbox{\sc Among}}
\newcommand{\mysum}{\mbox{\sc MySum}}
\newcommand{\amongseq}{\mbox{\sc AmongSeq}}
\newcommand{\atmost}{\mbox{\sc AtMost}}
\newcommand{\atleast}{\mbox{\sc AtLeast}}
\newcommand{\element}{\mbox{\sc Element}}
\newcommand{\gcc}{\mbox{\sc Gcc}}
\newcommand{\gsc}{\mbox{\sc Gsc}}
\newcommand{\contiguity}{\mbox{\sc Contiguity}}
\newcommand{\PRECEDENCE}{\mbox{\sc Precedence}}
\newcommand{\assignnvalues}{\mbox{\sc Assign\&NValues}}
\newcommand{\linksettobooleans}{\mbox{\sc LinkSet2Booleans}}
\newcommand{\domain}{\mbox{\sc Domain}}
\newcommand{\symalldiff}{\mbox{\sc SymAllDiff}}
\newcommand{\alldiff}{\mbox{\sc AllDiff}}

\newcommand{\slidingsum}{\mbox{\sc SlidingSum}}
\newcommand{\MaxIndex}{\mbox{\sc MaxIndex}}
\newcommand{\REGULAR}{\mbox{\sc Regular}}
\newcommand{\regular}{\mbox{\sc Regular}}
\newcommand{\precedence}{\mbox{\sc Precedence}}
\newcommand{\STRETCH}{\mbox{\sc Stretch}}
\newcommand{\SLIDEOR}{\mbox{\sc SlideOr}}
\newcommand{\NAE}{\mbox{\sc NotAllEqual}}
\newcommand{\mytheta}{\mbox{$\theta_1$}}
\newcommand{\mysigma}{\mbox{$\sigma_2$}}
\newcommand{\mysigmatwo}{\mbox{$\sigma_1$}}

\newcommand{\todo}[1]{{\tt (... #1 ...)}}
\newcommand{\myOmit}[1]{}
\newcommand{\nina}[1]{#1}
\newcommand{\ninacp}[1]{#1}

\newcommand{\dpsb}{DPSB}

\section{Introduction}

Thirty years ago, new constraint satisfaction algorithms
were often benchmarked on a small range of problems like
the zebra problem (which is unrepresentative of many problems as
it has an unique solution) or the $n$-queens problem
(which is also unrepresentative of many problems as
it has many solutions for large $n$), 
Over the next decade, experimental practice improved a little once it became the norm to
benchmark on hard, random problems (see, for example,
\cite{mitchell-hard-easy,SAT-phase,kirkpatrick1,gw-ecai94,gmpwcp95,pub702,achlioptas1,mpswcp98,ghpwaaai99}.
But random problems (whilst they can be hard to solve
if we generate them at some satisfiability phase
boundary) still don't represent many 
real world problems actually met in practice. 
There are structures common in real world problems 
that aren't met with any great frequency in simple random 
ensembles (see, for instance, \cite{isai95,wijcai99,random,wijcai2001}).

We are not that surprised our paper about CSPLib in the
CP-1999 proceedings \cite{csplib} attracted a lot 
of citations subsequently. It has over citations 300 on Google Scholar
today, putting it in the top three scientific papers either of us have
ever written. This is not because of any great scientific
merit or insights it contains. It simply reflects that our
intuitions two decades ago were correct. The constraint programming
community needed to benchmark on a larger and more representative
class of problems. And hundreds of researchers have 
taken advantage of this benchmarking resource subsequently. 

\section{A Very Brief History of CSPLib 
}

It seemed very natural to propose a benchmark library for constraints, following other areas of research such as Theorem Proving's TPTP \cite{Sut10} and Operations Research's OR-Lib \cite{beasley1990or}.  Seeing the good and bad points of other libraries also helped inform our opinions of what a good library should look like. Indeed, one of us (Ian) had engaged in an argument in the Journal of Heuristics on the pitfalls of benchmark libraries \cite{Gent1998,Falkenauer1998,Gent1999}. 
Toby started leading discussions in 1998 on how best to set up CSPLib for example giving presentations on what we were thinking. The value of this can be seen by a change of mind following a comment from Barbara Smith in one of those sessions. We had previously thought we would come up with a formal language to express problems in, but Barbara argued that problems should be described in English. While this inevitably makes it harder to run the problems, modelling is such a critical part of constraint solving that any formal expression, no matter how high-level, may compromise finding the best model. Barbara persuaded us, except that we changed `English' to `Natural Language', to allow for submissions in other languages. 

After deciding on the details we set up a simple website and included some initial problems. The next important phase was to publicise the library and also solicit contributions of problems. The site went public in 1999 and as part of this we decided to write a paper and submit it to CP 1999.  As well as the paper submission we put the full version online as a technical report \cite{csplib}\footnote{The technical report version is available at \url{https://tinyurl.com/CSPLib-html}}. Strange as it may seem for what is now a citation classic, our paper was in fact \emph{not} accepted in full.  We saw the logic that our paper was perhaps not a significant research contribution.  Instead, we were asked (and of course agreed) to produce a two page version for the proceedings and a poster for the conference itself.  History does seem to have confirmed the editor's words that poster papers were intended  `to showcase yet more interesting directions' \cite{DBLP:conf/cp/1999}.

An important part of the history of CSPLib has been that the editors pass on the baton of the website to avoid stagnation. As well as new ideas, new editors of the website can have more enthusiasm and perhaps better contacts with new generations of researchers to encourage submissions from.  Since foundation the leadership of CSPLib has changed twice. First, we handed over to Ian Miguel and Brahim Hnich, and later they in turn passed it to Chris Jefferson and \"Ozg\"ur Akg\"un (with some coding and maintenance help by Bilal Syed Hussain).

\section{Design considerations}

There are a number of desiderata for any benchmark
library. 
\begin{description}
\item[Easy to find:] 
Just go to CSPLib.org. It's hard
to imagine a simpler or more apt URL.
\item[Large and diverse:] 
the library needs to be diverse, so we don't 
over-fit. It needs to grow continuously,
again to prevent over-fitting. 
And it needs to contain a mix of problems
of varying difficulty: hard and easy problems,
open and solved problems, and problems 
of a wide variety that . 
\item[Easy to use:] if there is one area
in which CSPLib fails, it is this. 
You can't simply read in problems with a 
provided parser as you can with, say, SATLib or
MIPLib. You
have to understand the natural language
description, work out a good model, and 
implement this yourself. But given that we didn't 
want to commit users to a particular representation,
it's hard to imagine how we could have avoided this? 
\item[Ever growing:]
Benchmark libraries like CSPLib should always be growing.  
Unfortunately, the last problem submitted to CSPLib
was about a year ago. Perhaps you can fix this? 

\end{description}

\section{An ongoing concern}

The challenge that CSPLib always had (which we suspect
is probably true of almost every other benchmark library too)
is that, whilst people often use the benchmarks, few
contribute to it. In part, this is structural. What credit
do you get for going to the effort of submitting problems? 
This is not an easy problem to fix. 
Economists might perhaps suggest 
we need to design a market in which people had
an incentive to submit. 
 For example, the current maintainers introduced an incentive that might help.  Each problem page in CSPLib provides bibtex to give credit to the proposer of each problem.  For example, the largest numbered problem at time of writing is Problem 133, the Knapsack problem, which we hereby cite \cite{csplib:prob133}. Another activity has been CSPLib hackathons at CP conferences, encouraging people to sit in a room together to write new entries.

\section*{Acknowledgments}

We thank all those who helped us set up CSPLib, especially Barbara Smith and Bart Selman. We thank all subsequent editors of CSPLib, Ian Miguel, Brahim Hnich, Chris Jefferson and \"Ozg\"ur
Akg\"un, and also Bilal Syed Hussain for his technical work on the site.
Most importantly, we thank all the contributors to CSPLib over the past twenty years.
Toby Walsh is supported by
the European Research Council under the Horizon 2020 Programme via AMPLify
670077.

\bibliographystyle{plain}
\bibliography{/Users/tw/Documents/biblio/a-z,/Users/tw/Documents/biblio/a-z2,/Users/tw/Documents/biblio/pub,/Users/tw/Documents/biblio/pub2,csplib-new}


\end{document}